\documentclass{article}

\usepackage[final]{corl_2024}

\usepackage[utf8]{inputenc} 
\usepackage[T1]{fontenc}    
\usepackage{hyperref}       
\usepackage{url}            
\usepackage{booktabs}       
\usepackage{amsfonts}       
\usepackage{nicefrac}       
\usepackage{microtype}      
\usepackage{xcolor}         
\usepackage{import}   
\usepackage{lipsum}             
\usepackage{amssymb}
\usepackage{acronym}
\usepackage{algorithm}
\usepackage{algpseudocode}
\usepackage{multirow}
\usepackage{caption}
\usepackage{subcaption}
\usepackage{wrapfig}
\usepackage{multicol}
\usepackage{graphicx}
\usepackage{amsmath}
\usepackage{soul}
\graphicspath{{./img/}}

\newcommand{\phil}{\color{blue}}

\title{Iterative Batch
Reinforcement Learning via
Safe Diversified
Model-based Policy Search}

\author{
Amna Najib, Stefan Depeweg, Phillip Swazinna \\
Siemens AG \\
 \texttt{\{amna.najib, stefan.depeweg, phillip.swazinna\}@siemens.com} \\
}

\begin{document}

\maketitle

\begin{abstract}
Batch reinforcement learning enables policy learning without direct interaction with the environment during training, relying exclusively on previously collected sets of interactions. This approach is, therefore, well-suited for high-risk and cost-intensive applications, such as industrial control. Learned policies are commonly restricted to act in a similar fashion as observed in the batch.  In a real-world scenario, learned policies are deployed in the industrial system,  inevitably leading to the collection of new data that can subsequently be added to the existing recording. The process of learning and deployment can thus take place multiple times throughout the lifespan of a system. In this work, we propose to exploit this iterative nature of applying offline reinforcement learning to guide learned policies towards efficient and informative data collection during deployment, leading to continuous improvement of learned policies while remaining within the support of collected data. We present an algorithmic methodology for iterative batch reinforcement learning based on ensemble-based model-based policy search, augmented with safety and, importantly, a diversity criterion. 
\end{abstract}

\section{Introduction}

The objective of batch (or offline) reinforcement learning (RL) is to extract the best possible behavior out of  existing data, called a batch, without any learning during deployment. This implies that learning is more successful if the  initial data is diverse or collected through the deployment of an expert agent. In a real setting, the initial batch is prone to limitations (low data coverage, low reward actions, etc.), forming a challenge in learning for real-world applications.  This challenge of limited information calls for safety mechanisms, such as regularization, to ensure reliable performance of the agent \cite{swazinna2021overcoming,yu2021combo}.

In many industrial setups, the application of offline reinforcement learning is not a one-time process  but iterative. After an RL agent is trained and deployed on the system, a new set of recordings becomes available.  
The principal contribution of our work relies on the formulation of iterative batch reinforcement learning ({\textbf{IBRL}), a novel framework to iteratively refine the initial data batch and improve learned policies after each new batch collection, without dropping performance due to overly adventurous exploration. In every iteration, we seek to improve the data coverage by deploying a set of policies, that we previously trained to be diverse, i.e. that act in a variety of ways to explore, without compromising the rewards too much. The proposed IBRL algorithms adhere to safety constraints by restricting the state or action space of the learned policies depending on the data support. Through experiments in an illustrative 2D environment, as well as on the Industrial Benchmark \cite{hein2017benchmark}, we demonstrate the improved exploration capability and resulting improved performance of our approach, all while maintaining safety considerations and not underperforming the behavioral.
Conceptually, our work is most closely related to \cite{matsushima2020deployment, hu2023sample, Zhang2023}, however these works do not address the combination of diversity and safety, or focus solely on the iterative process without incorporating an exploration incentive.

\section{Related Work}
\textbf{Offline RL}: Early works in the so-called ``batch RL'' setting include \cite{ernst2005approximate,riedmiller2005neural,riedmiller2009reinforcement,lange2012batch}, as well  as more recently \cite{hein2016reinforcement,depeweg2016learning,hein2018interpretable,depeweg2018decomposition}. While these works focused on reinforcement learning in the batch setting, a key distinction to later proposed offline methods is that they mostly assume the initial batch to be collected under uniform random actions, allowing a relatively well explored environment. While the batch RL setting is certainly closer to the requirements imposed by real-world deployments of RL algorithms than commonly popular online algorithms such as \cite{silver2014deterministic,schulman2017proximal,haarnoja2018soft}, it is still not exactly what industry practitioners need most of the time. Algorithms such as \cite{DBLP:journals/corr/abs-1812-02900,kumar2019stabilizing,DBLP:journals/corr/abs-1911-11361,kumar2020conservative,DBLP:journals/corr/abs-2106-06860,DBLP:journals/corr/abs-2110-06169} employ various regularization techniques to keep the trained policies in regions of the environment where the models are sufficiently accurate. Most common techniques include behavior regularization \cite{DBLP:journals/corr/abs-2002-08396, kumar2019stabilizing} and uncertainty-based regularization \cite{jin2022pessimism, DBLP:journals/corr/abs-2110-08695}. 
In offline RL, model-based RL appears to have an edge over model-free methods, which enjoy better asymptotic performance but have generally been attributed lower data efficiency \cite{deisenroth2011pilco,DBLP:journals/corr/abs-2110-06169,nagabandi2018neural}. Methods such as \cite{depeweg2018decomposition,swazinna2021overcoming,yu2020mopo,kidambi2020morel} have thus developed ways to extend action divergence regularizations to the model-based setting. \textbf{Offline-to-Online}: Some of the early batch RL works mentioned above as well as \cite{matsushima2020deployment, li2023proto} introduced the idea of the growing batch setting, a problem in between offline and online learning where one is constrained to only deploy a limited number of times to the real system to collect new data. While the idea is appealing, our approach differs significantly since their almost unconstrained exploration can be an issue due to safety constraints or requirements on the minimal performance. Other works that explored online finetuning of offline pretrained policies without the deployment constraint exhibit similar issues due to their exploration strategies \cite{mark2022finetuning, DBLP:journals/corr/abs-2110-06169, DBLP:journals/corr/abs-2006-09359, DBLP:journals/corr/abs-2107-00591, hu2023iteratively}.
\textbf{Diversity}: Intrinsically motivated agents have been introduced in \cite{schmidhuber1991possibility}. Agents have since been found to perform effective exploration using intrinsic reward objectives like curiosity or diversity and have been studied e.g. in \cite{pathak2017curiosity, DBLP:journals/corr/abs-1802-06070, DBLP:journals/corr/abs-1802-04564}. Previous work has also explored the effect of collecting initial diverse data that is further used for offline learning in several downstream tasks \cite{DBLP:journals/corr/abs-2201-13425, DBLP:journals/corr/abs-2201-11861, hong2018diversity}, as well recently \cite{parker2020effective, kumar2020one}.

\section{Method}
\begin{figure}[t]
\centering
\includegraphics[width=0.7\textwidth]{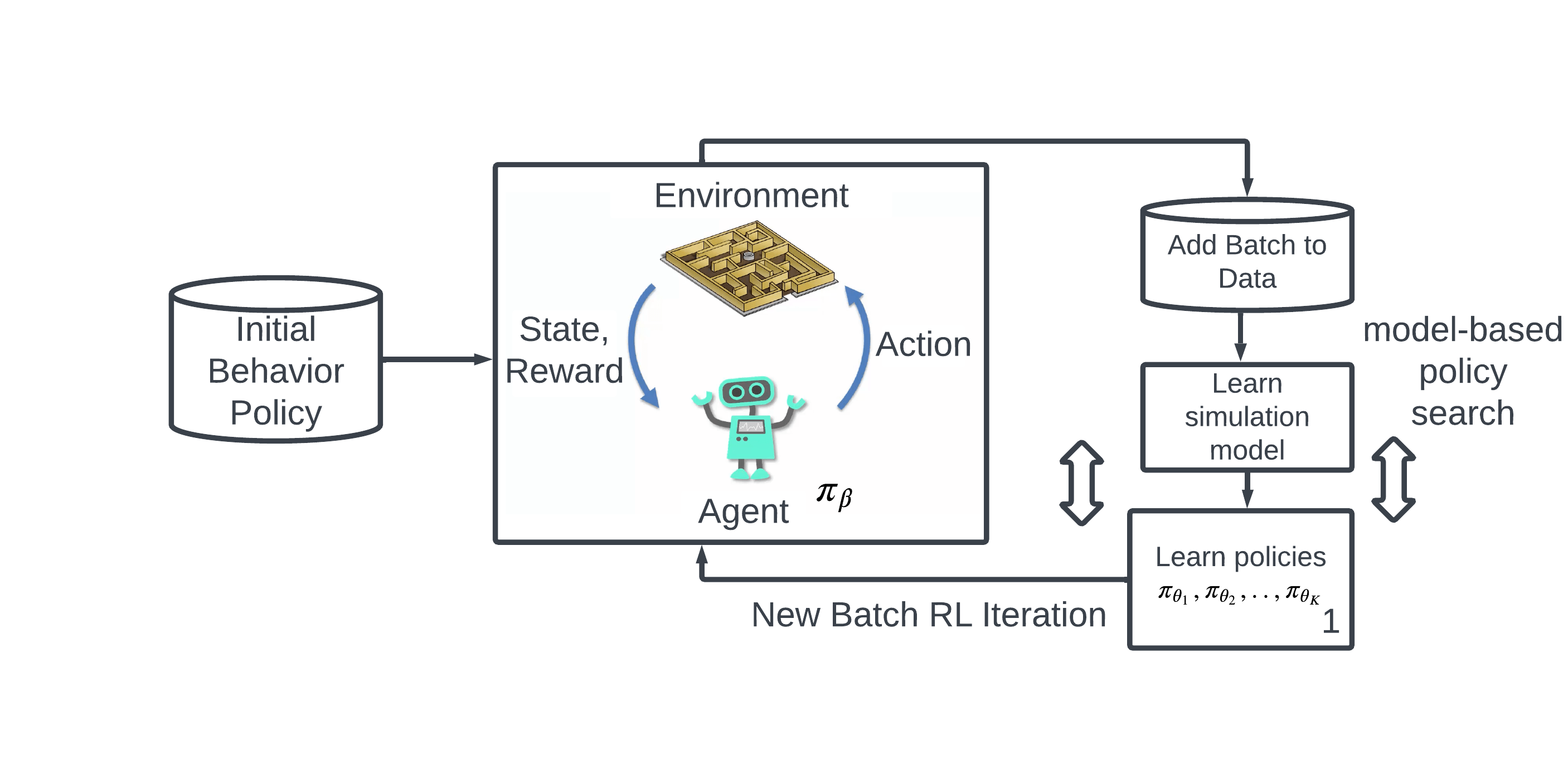}
\caption{Illustration of iterative model-based policy search.}
\label{fig:imbps}
\end{figure}

Offline reinforcement learning uses a fixed dataset $\mathcal{D}$ collected by one or more  behavior policies $\pi_{\beta}$ to learn a policy $\pi$.   Once trained, the agent's policy is fixed and no further learning occurs during deployment. To ensure safety, the learned policy is often constrained to avoid significant deviations from the original behavior policies used to generate the data.

One  approach for learning a policy is a model-based policy search. This is a two-step process. First, a simulation model is learned from the available data. In the second step, one or a set of policies are optimized using virtual rollouts. For policy search, we seek to minimize the loss function
\begin{equation}
    L(\theta) = - \frac{1}{N} \frac{1}{K} \frac{1}{H} \sum_{k=1}^{K} \sum_{s_{k, 1} \sim D} \sum_{t=1}^{H} \gamma^t e(s_{k, t}, a_{k, t}) \;,
\end{equation}
where K is the number of policies, $e(s_{k, t}, a_{k, t})$ the reward received while taking action $a_{k, t}$ at state $s_{k, t}$ and $\theta=\theta_1,\ldots,\theta_K$ the parameters of the policies. 
We aim to maximize the accumulated reward of trajectories generated by following the K learned policies. Every trajectory is formalized as
\begin{equation} 
\begin{split}
T_k & = \{s_{k, 1}, a_{k, 1}, s_{k, 2}, a_{k, 2} .. s_{k, H-1}, a_{k, H-1}, s_{k, H}\}  \\
 & = \{ s_{k, i} \mid s_{k, i} = f(s_{k, i-1}, a_{k, i-1}; \eta), a_{k, i-1} = \pi(s_{k, i-1}; \theta_k), 1 < i < H + 1, s_{k, 1} \sim D\} 
\end{split}
\end{equation}
where $f(\cdot; \eta)$ denotes the learned transition model.

In \textbf{iterative offline reinforcement} learning the offline RL setting is repeated at different times over the lifetime of a system. The collected data in every iteration of deployment is added to the batch and further used to refine the training of the policy. We can apply the method of model-based policy search to the iterative offline setting: every time a new batch of data becomes available, we update the transition model and then redo the policy training and deployment. This will serve as the backbone of our proposed algorithm, which we illustrate in Figure  \ref{fig:imbps}. We note that alternative approaches, such as a model-free approach are viable as well. We choose model-based policy search, because it tends to work well in practice, can be easily extended (e.g. via uncertainty modeling), and allows automatic differentiation in continuous state, action, and reward spaces.

Importantly, we argue that two components are highly beneficial for model-based policy search in the iterative batch scenario: \textbf{safety} and \textbf{diversity}. The former is ubiquitous in all real-world applications and should safeguard against exploiting model inaccuracies due to limited data. The latter should exploit the iterative nature and improve information gain over each iteration. We hypothesize that an ideal algorithm for iterative batch RL is safe but diverse.

\subsection{Safety}
We formulate three approaches towards realizing safety in policy search. Safety may be used (i) as an additional objective in the loss function, (ii) as a soft constraint, or finally, (iii) it may also be possible to constrain the policy directly as part of its architecture.
\paragraph{Safety as an objective}

Safety is injected as an explicit objective in the loss function to balance between high rewards and not deviating from the behavior policy. It was used in previous research work including \cite{swazinna2022user, brandfonbrener2021offline, DBLP:journals/corr/abs-1911-11361}. At first, a behavior policy is learned in a supervised manner from the fixed initial dataset as denoted in Eq. (\ref{behavLearn}). Afterwards, for every state, the behavioral policy action and the new policy actions are predicted and used to define the deviation between the behavior policy and the learned policy. The deviation in the simplest case is a mean-squared error between both actions. If a distribution of actions is learned for both policies, the Kullback-Leibler divergence (KL divergence) could instead be used to assess the deviation.  In the MSE case we have:
\begin{equation} \label{behavLearn}
L(\phi) = \sum_{s, a \sim D} \lVert a - \pi_{\beta}(s;\phi) \rVert_2    \;.
\end{equation}
Then, the loss function including the reward maximization and safety objectives is defined as 
\begin{equation}  \label{safetyloss}
\begin{split}
 L(\theta) &= \frac{1}{K} \frac{1}{H} \sum_{k=1}^{K} \sum_{s_{k, 1} \sim D}   \sum_{t=1}^{H}   [- \gamma ^ t \lambda e(s_{k, t}, a_{k, t}) + (1- \lambda) p(a_{k, t})  \;, \\
p(a_{k, t}) &= \lVert \pi(s_{k, t};\theta_k) - \pi_{\beta}(s_{k, t};\phi) \rVert_2 \;.
\end{split}
\end{equation}
While this is the standard approach for safe offline RL it has one key drawback: Weighing safety against performance (and possibly diversity) in the form of a trade-off (in this case via the parameter $\lambda$) appears counter-intuitive for safety-critical applications.

\paragraph{Safety as a soft constraint}
An alternative approach is to specify a loss term that is flat inside a safe region and then provides a large loss value outside, thereby implementing a differentiable constraint. In contrast to the aforementioned objective-based approach, here the safety term is not weighted against the objective but instead is effectively restricting the allowed solution space of the policy.

We note that in iterative batch RL, the transitions in the batch may be generated by the execution of different policies. To that end, we propose to approximate the behavior policy using a probabilistic approach.
We assume that the learnt behavior policy is a Gaussian distribution with mean $\mu_\beta$ and diagonal covariance matrix $\Sigma_\beta$, $\pi_{\beta}(s;\phi) \sim  \mathcal{N}(\mu_\beta,\,\Sigma_\beta)$. $\Sigma_\beta$ reflects the aleatoric uncertainty in the model. A prediction has a high aleatoric uncertainty in the case of diverse actions present in the behavior data. 
As a measure of safety we consider how likely an action is under the behavior policy. We make 3 considerations for the proposed metric: to be normalized between 0 and 1, to be sensitive to low likelihoods in certain action dimensions and lastly to be more permissible in  situations where diverse actions were executed. To that end we use the negative unnormalized likelihood and compute the geometric mean over the action space: 
\begin{equation}
G(S, A) = -[\prod_{d} \exp\left(-\frac{1}{2} (a - \mu_\beta(s; \phi))^T \Sigma_\beta(s; \phi)^{-1} (a - \mu_\beta(s; \phi))\right) ]^{\frac{1}{d}}    \;, \label{eq:geo_mean} 
\end{equation}
where $d$ is the dimensionality of the actions. 
 To enforce safety during training of policy $\pi$ we can use thresholding: 
\begin{equation} 
L_S(\theta)  = \alpha_s \frac{1}{K} \frac{1}{H} \sum_{k=1}^{K}   \sum_{t=1}^{H} \max(G(S, A) + \delta,0) \quad \mathrm{with} \quad S \sim f(\cdot;\eta), A \sim \pi(S;\theta) \;,\label{eq:soft_constr}
\end{equation} 
where $\delta \in [0, 1]$ represents the safety threshold and controls the permissibility of the training. Any deviation of the actions represented by $G(S, A)$ lower than $-\delta$ is not penalized in the learning. To conclude, the loss in Eq. (\ref{eq:soft_constr}) implements a likelihood-based safety zone. 

\paragraph{Constrained Policy}
The direct approach to fulfill safety is arguably to directly constrain the expressiveness of the policy itself. In this scenario, we consider safety with respect to the state space and not relative to a behavior policy. We assume that professionals may have prior knowledge about safety ranges (bounds) of sensors. We start building on this assumption and limit all the states over all virtual rollouts for the different ensembles to lie within the predefined bounds. We inject this bound constraint in the policy specification. 

This approach is only applicable if actions $a$ affect a subset
of system variables $s'$ in a known linear way. Let $a_{\text{lower}}$ and $a_{\text{upper}}$ be the lower and upper bounds of actions. Further, let
$\pi(s_t;\theta)$ be designed such that only actions in this bound can be computed.  Let $B1$ and $B2$ be the lower and upper safety bound of the state. 
We then compute  the valid action range $a_{\text{min}},a_{\text{max}}$ such that $B1 < s'(t+1) < B2$.
We then define the constrained policy such that:
\begin{equation} \label{constrainedp}
 \pi_{\text{constr}}(s_t;\theta_k) = a_{\text{min}} + (a_{\text{max}} - a_{\text{min}}) * \frac{\pi_{\text{constr}}(s_t;\theta_k) - a_{\text{lower}}}{a_{\text{upper}} - a_{\text{lower}}}
\end{equation}
\subsection{Diversified Policy Search}

We define diversity as the ability to discover different or dissimilar state regions. This translates back to having a high entropy on the trajectory samples used for training and deployment. Given an ensemble of $K$ policies, we draw $K$ trajectory samples $T_1,\ldots,T_K$, where $T_k$  results from the interaction between the learned transition model $f(s, a;\eta)$, reward model $f(s, a;\omega)$ and  policy $\pi_{\theta_i}$ under the same starting state $s_1$. Let $D(\theta, \eta, \omega)$ denote the distances between all trajectories. The diversity-based exploration enforces the maximization of $D(\theta, \eta, \omega)$, by adding the diversity loss as an intrinsic motivation for exploration.

\begin{equation*} 
L_d(\eta, \omega, \theta) = -D(T_1, T_2, .., T_K)  = -D(\theta, \omega, \eta)
\end{equation*}
\paragraph{Diversity measures} \label{diversity_measures}
We wish for the diversity-based objective to be differentiable w.r.t $\theta_k$ \cite{4308316, bottou2018optimization}. Different distance metrics between trajectories were used in different communities \cite{Mangalam, Choi, Zhao}. The simplest distance metric is the pairwise distance between trajectories, also called lock-step Euclidean distance (LSED). LSED measures the spatial discrepancy between time-corresponding states over the rollout of different K ensemble policies. The distance between trajectory $T_i$ and $T_j$ is calculated as
\[ D(T_i, T_j)= \frac{1}{H} \sum_{t=1}^{H} \lVert s_{i, t} - s_{j, t} \lVert_2 \]
The pairwise distance between all trajectories induced by the different K policies is
\begin{equation} 
D = \frac{1}{K!}  \sum_{k=1}^{K} \sum_{k'=1, k'\neq k}^{K} D(T_k, T_{k'})  = \frac{1}{H} \frac{1}{K!}    \sum_{k=1}^{K} \sum_{k'=1, k'\neq k}^{K} \sum_{t=1}^{H} \lVert s_{k, t} - s_{k', t} \lVert_2
\end{equation}

 LSED diversity can suffer from outlier behavior. Due to the fact that diversity and reward maximization potentially act as conflicting objectives, LSED diversity can incentivize all policies but one to focus on cost minimization and learns one outlier policy to bring the diversity to a high value, by that fulfilling the tradeoff to balance these competing objectives. Using the L1 norm instead of the L2 norm reduces the outlier behavior to an extent.

An alternative approach is to instead use the minimum pairwise distance between trajectories (MinLSED) to mitigate the outlier behavior of LSED. The MinLSED diversity represents the minimum mean distance over the horizon of the pairwise trajectories and is formalized as
\begin{equation}
 D = \frac{1}{H} \min_{k'\neq k,k \in K}  D(T_k, T_{k'}) \;. \label{eq:minLSED}
\end{equation}
In this work we will utilize the MinLSED diversity specified in Eq. (\ref{eq:minLSED}). 
\subsection{Algorithm}\label{sec:method}
\begin{wrapfigure}{r}{0.44\textwidth}
\vspace{-25pt}
\begin{minipage}{0.44\textwidth}
\begin{algorithm}[H]
\begin{algorithmic}
    \State \textbf{Input:} Data \( \mathcal{D} \), behavior policy \( \pi_\beta(\mathbf{s}; \phi) \)
    \State Train transition model \( f(s_t, a_t; \eta) \) on \( \mathcal{D} \)
    \State Train reward function \( f(\mathbf{s},a; \omega) \) on \( \mathcal{D} \)
    
    \While{not converged}
        \State \(s_{k,1} \sim \mathcal{D}\)
        \For{k=1,\ldots,K}
            \For{t=2,\ldots,H}
                \State $\mu_\beta(s_{k,t}),$
                \State $\Sigma_\beta(s_{k,t}) = \pi_\beta(s_{k,t}; \phi)$
                \State $a_{k,t} = \pi(s_{k,t};\theta_k)$
                \State $s_{k,t+1} = f(s_{k,t}, a_{k,t}; \eta)$
                \State $r_{k,t}  =  f(s_{k,t},a_{k,t}; \omega)$
                \State  $R_k += \gamma^t r_{k,t}$ 
                \State $ s_{k,t} = s_{k,t+1}$
            \EndFor
        \EndFor
        \State Compute  Eq. (\ref{eq:minLSED})  and  Eq. (\ref{eq:soft_constr})
        \State Train \(\pi(.;\theta_1)\ldots,\pi(.;\theta_k)\) on Eq. (\ref{eq:Loss_algo})
    \EndWhile
\end{algorithmic}
\caption{Safe Diversified model-based policy search (Soft Constraint)}
\label{algo:mbps1}
\end{algorithm}
\end{minipage}
\end{wrapfigure}

We will show here one instantiation of a possible loss function using a soft-constrained policy and minLSD diversity.
\begin{align}
L&\theta) =  -\frac{1}{NKH} \sum_{k=1}^{K} \sum_{\substack{s_{k,1} \sim \mathcal{D}}} \sum_{t=1}^{H} \gamma^t e(s_{k,t}, a_{k,t}) \nonumber \\
& + \alpha_s \frac{1}{KH} \sum_{k=1}^{K} \sum_{t=1}^{H} \max(G(s_{k,t}, \pi(s_{k,t}; \theta_k)) + \delta, 0) \nonumber \\
& - \alpha_d \frac{1}{H} \min_{\substack{k' \neq k, k \in K}} D(T_k, T_{k'}) \label{eq:Loss_algo}
\end{align}
where we note that in Eq. (\ref{eq:Loss_algo}) by default we reduce the scope of the diversity term to exclude one (the first) policy. The reasoning is to have always one policy available that is only influenced by the reward and the safety. All policies are used for data generation, while only the first is used for evaluating the performance.

The aforementioned instantiation of the training algorithm for one batch iteration is shown in Algorithm \ref{algo:mbps1}. 
\vspace{0.2cm}
 \endwrapfigure
This process is repeated whenever a new batch arrives  
after execution of the previous trained set of policies, with the batch appended to the existing data set, following the schema illustrated in Figure \ref{fig:imbps}.

\section{Experiments}
We investigate the advantage of the proposed iterative batch reinforcement learning framework in improving learned policies. We seek to investigate the supplementary advantage of incorporating diversity in the process. Finally, we investigate how different forms of safety impact the diversity objective. We evaluate our methods on a 2D grid environment and the industrial benchmark \cite{hein2017benchmark}.

\subsection{2D Grid Environment: Single Iteration}
The 2D grid environment is a simplistic benchmark illustrating common navigation tasks. The state (x, y) represents the position of an agent in space. The agent is rewarded according to its position, following a Gaussian distribution with mean $\mu = (3, 6)^T$ and diagonal covariance matrix $\Sigma$ with standard deviation vector $(1.5, 1.5)^T$. Concretely, the reward of an agent at state $s_t$ is the likelihood of the reward Gaussian distribution: $r(s_{k,t}) = \frac{1}{(2\pi)^\frac{3}{2}\lvert \Sigma \rvert^\frac{1}{2}} e^{-\frac{1}{2}(s_{k, t} - \mu)^T\Sigma^{-1}(s_{k, t} - \mu)}$. See Figure \ref{fig:2Denv} for a visualization of the reward distribution. We gather the initial data by deploying a behavior policy that navigates to one of the behavior goals denoted in Figure \ref{fig:2Denv} and fixed to the positions $(2.5, 2.5)^T$ and $(7.5, 7.5)^T$. The agent navigates to the goal closest to its current position. Additionally, the behavior policy is augmented with 10\% uniform random actions.
\begin{figure}[t]
\centering

\includegraphics[width=0.45\linewidth]{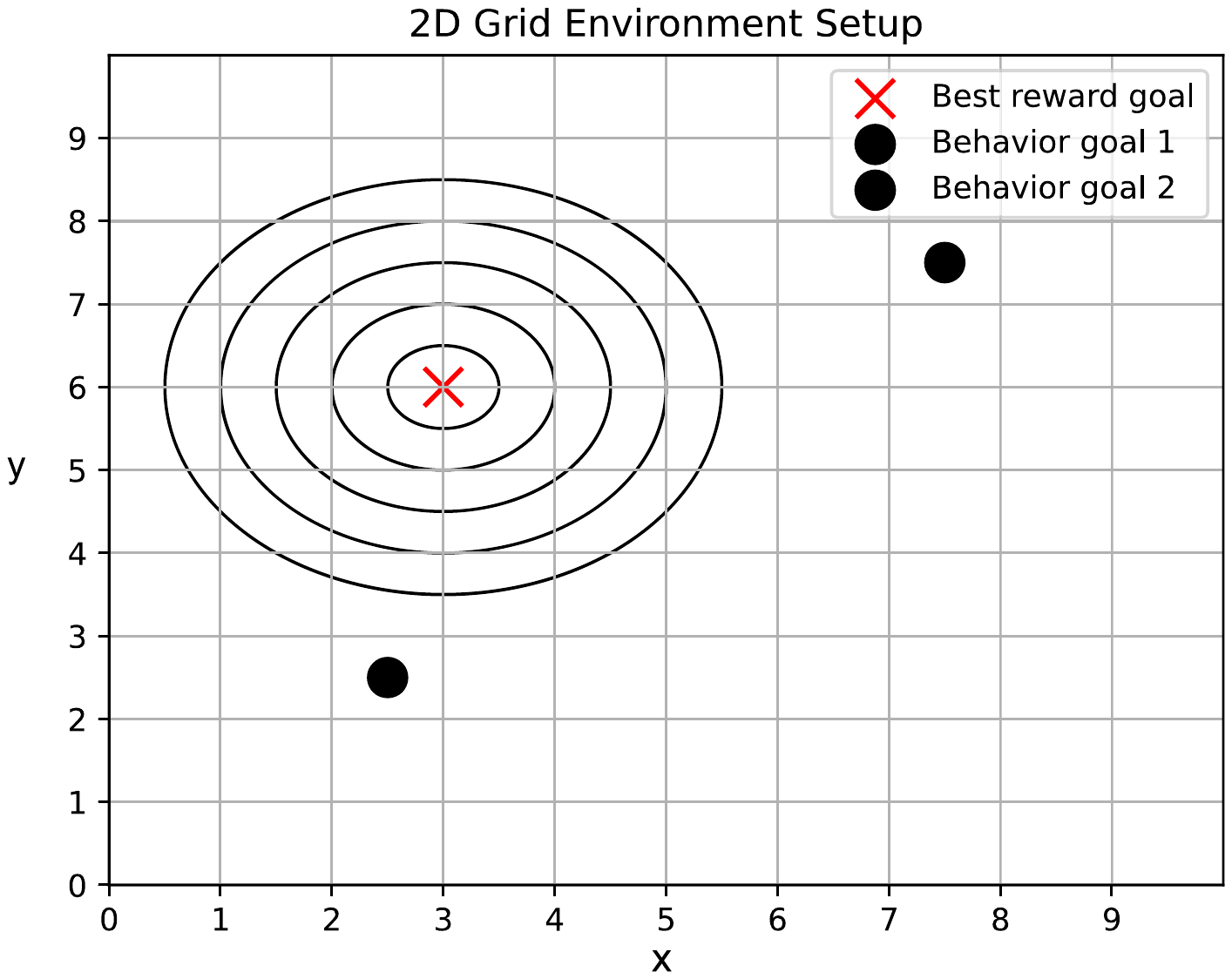}
  \caption{2D grid environment: The behavior policy guides agent towards the nearest behavior goal. The best reward goal represents the state with the highest reward. Reward decreases according to Gaussian distribution represented by circle lines.}
  \label{fig:2Denv}
\end{figure}

\begin{figure}[t]
\centering
\includegraphics[width=0.16\linewidth]{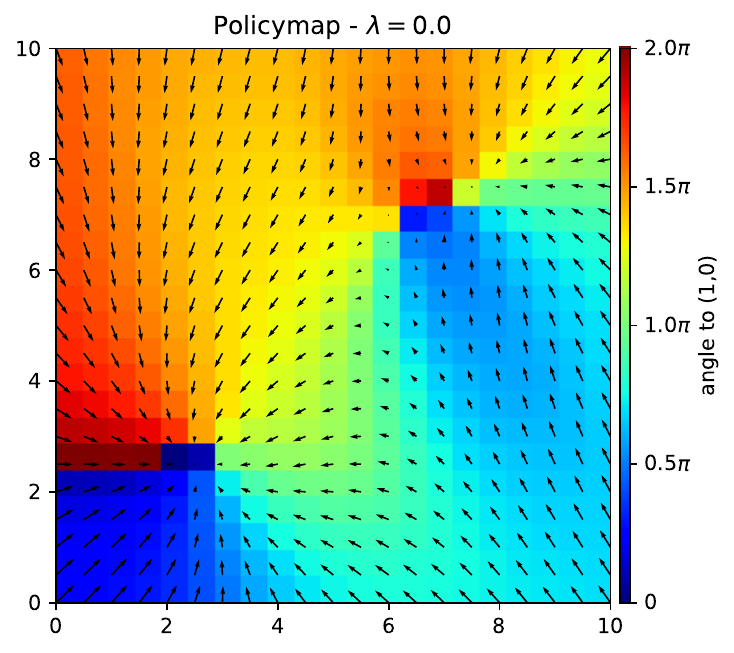}
\includegraphics[width=0.16\linewidth]{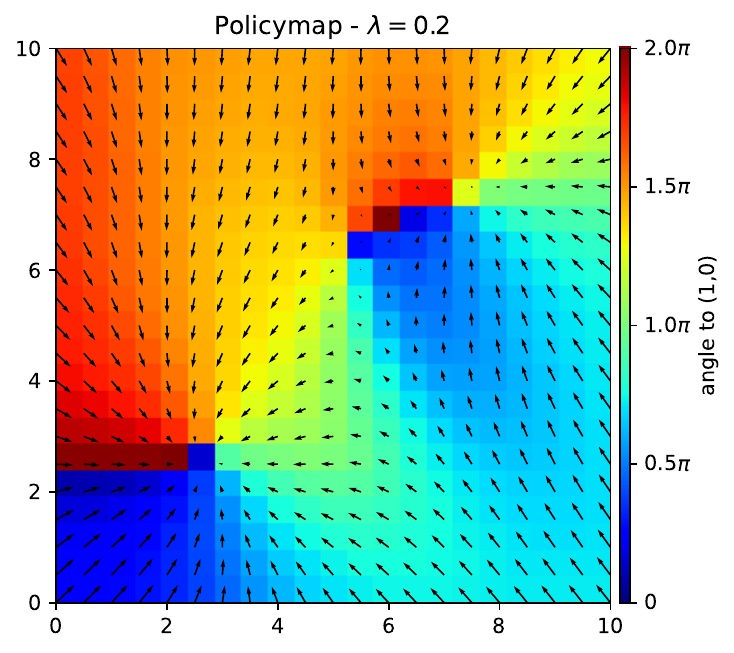}
\includegraphics[width=0.16\linewidth]{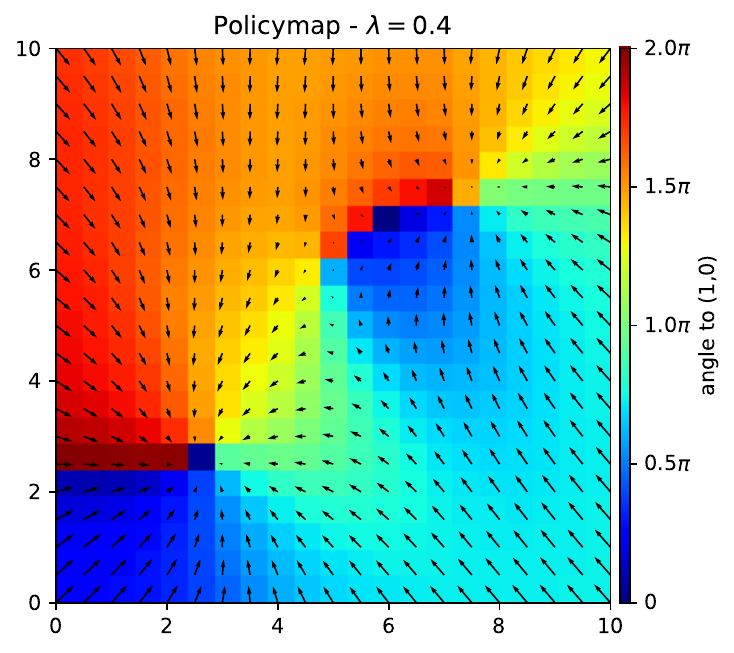}
\includegraphics[width=0.16\linewidth]{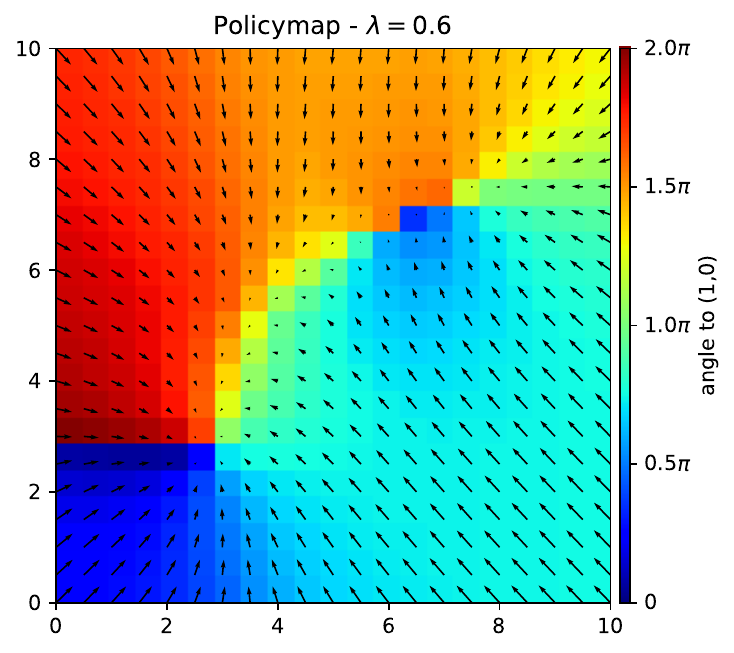}
\includegraphics[width=0.16\linewidth]{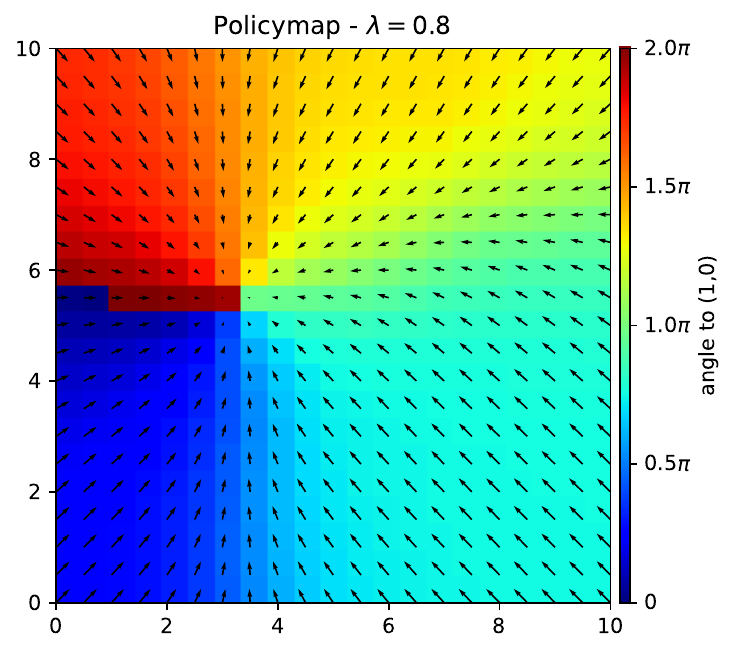}
\includegraphics[width=0.16\linewidth]{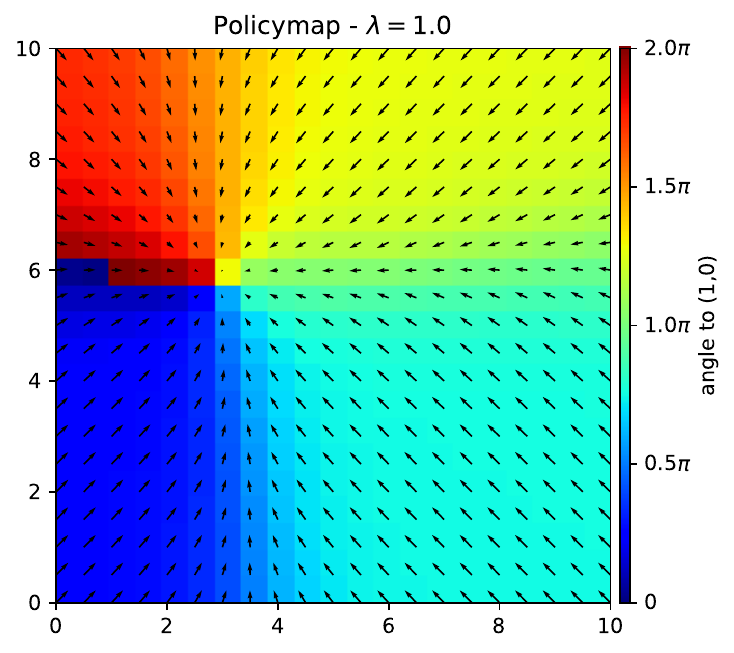}

  \caption{Policy maps for different $\lambda$ values,  colors illustrate  action directions in every cell of the grid. For $\lambda = 0.0$, the policy imitates behavior policy by navigating to the closest behavior goal. With increasing $\lambda$ values, the policy moves slowly towards the reward goal.}
  \label{fig:2DEnvVis}
\end{figure}

For reference, we first train a single policy without diversity for different values of $\lambda$ for one single iteration. Figure \ref{fig:2DEnvVis} represents the policy maps of actions taken by policies learned with increasing $\lambda$ values in the 2D grid environment. With $\lambda = 0.0 $, the objective is reduced to mimicking the behavior policy. In the case of $\lambda = 0.4$, the policy predominantly adheres to the behavior policy with slight adjustments in the direction of certain actions. Nevertheless, convergence towards both behavior poles is still observable. 
This validates our further assumption of fixing $\lambda$ at $0.4$. \newline

\begin{figure}[t]
\minipage{0.17\textwidth}
  \includegraphics[width=\linewidth]{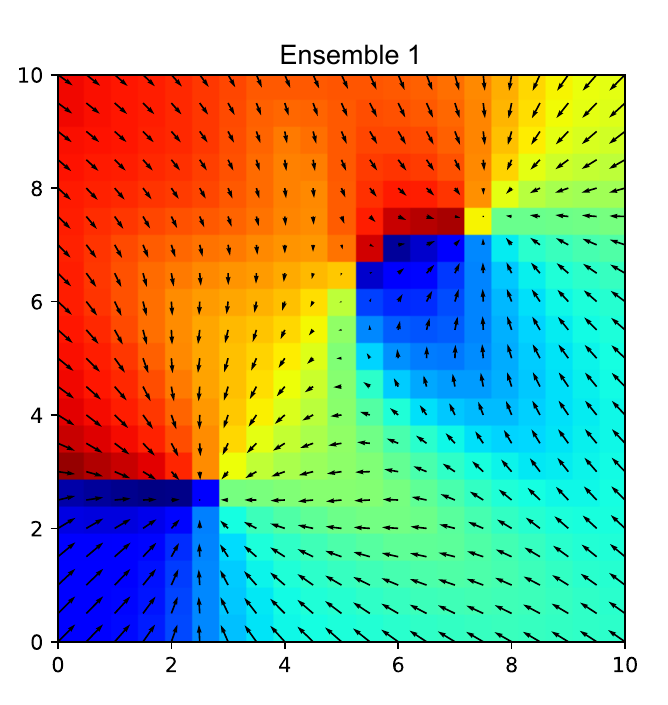}
  \captionsetup{justification=centering}
  \caption*{Ensemble 1}\label{fig:ens1nodiv}
\endminipage\hfill
\minipage{0.17\textwidth}
  \includegraphics[width=\linewidth]{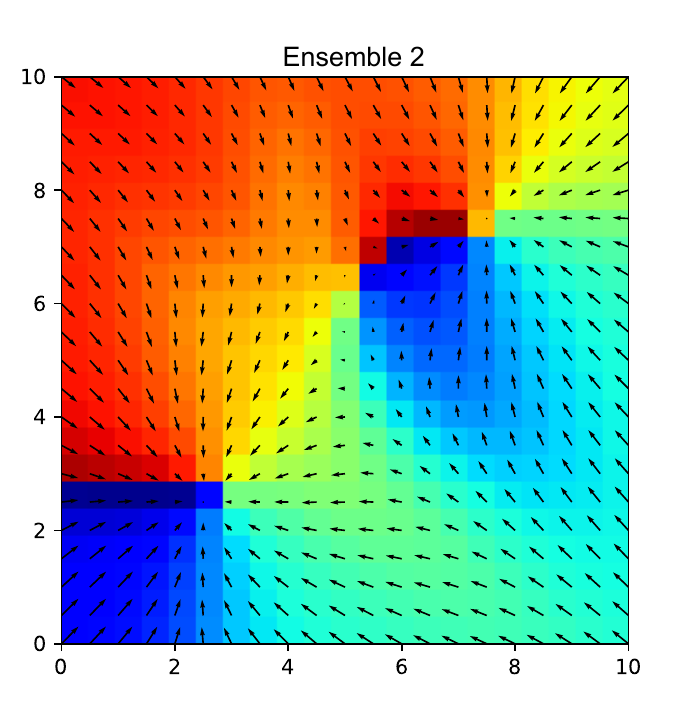}
  \caption*{Ensemble 2}\label{fig:ens2nodiv}
\endminipage\hfill
\minipage{0.17\textwidth}%
  \includegraphics[width=\linewidth]{img/ens2_div0.pdf}
  \caption*{Ensemble 3}\label{fig:ens3nodiv}
\endminipage\hfill
\minipage{0.17\textwidth}%
  \includegraphics[width=\linewidth]{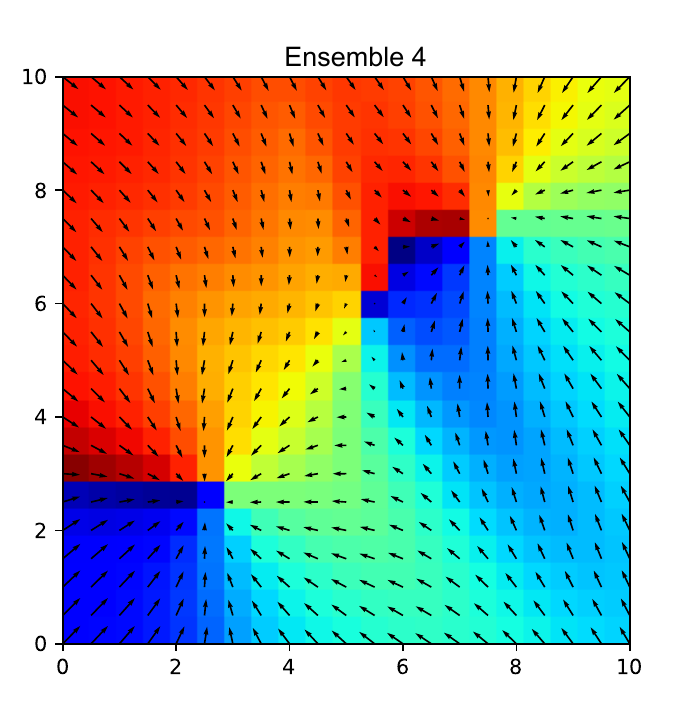}
  \caption*{Ensemble 4}\label{fig:ens4nodiv}
\endminipage
\\
\minipage{0.17\textwidth}
  \includegraphics[width=\linewidth]{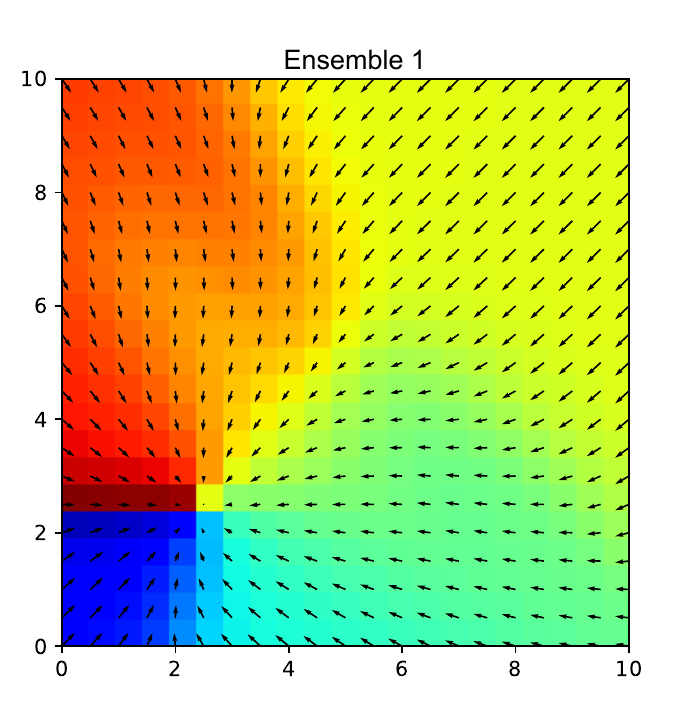}
  \captionsetup{justification=centering}
  \caption*{Ensemble 1}\label{fig:ens1div}
\endminipage\hfill
\minipage{0.17\textwidth}
  \includegraphics[width=\linewidth]{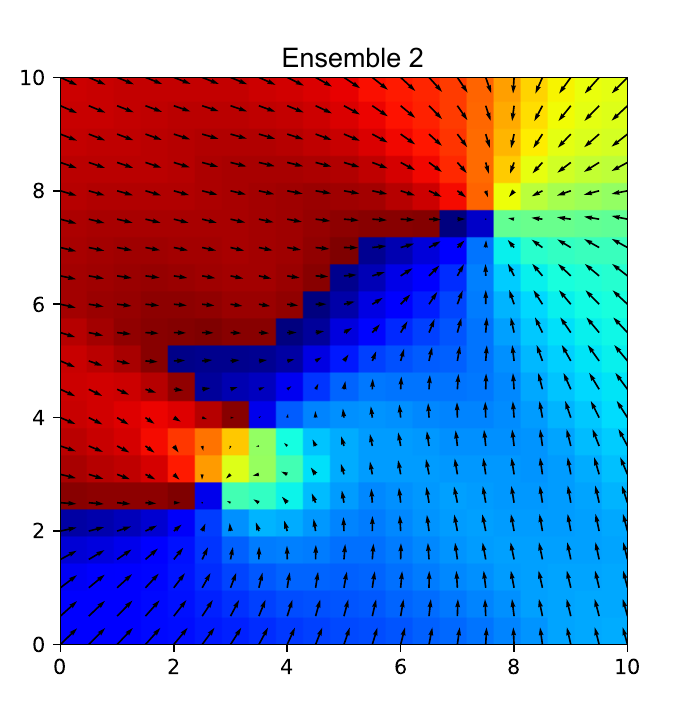}
  \caption*{Ensemble 2}\label{fig:ens2div}
\endminipage\hfill
\minipage{0.17\textwidth}%
  \includegraphics[width=\linewidth]{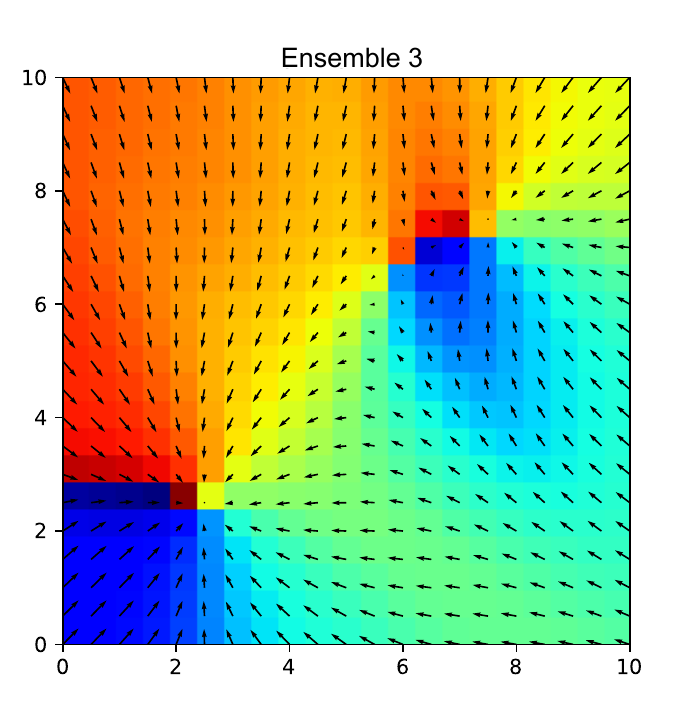}
  \caption*{Ensemble 3}\label{fig:ens3div}
\endminipage\hfill
\minipage{0.17\textwidth}%
  \includegraphics[width=\linewidth]{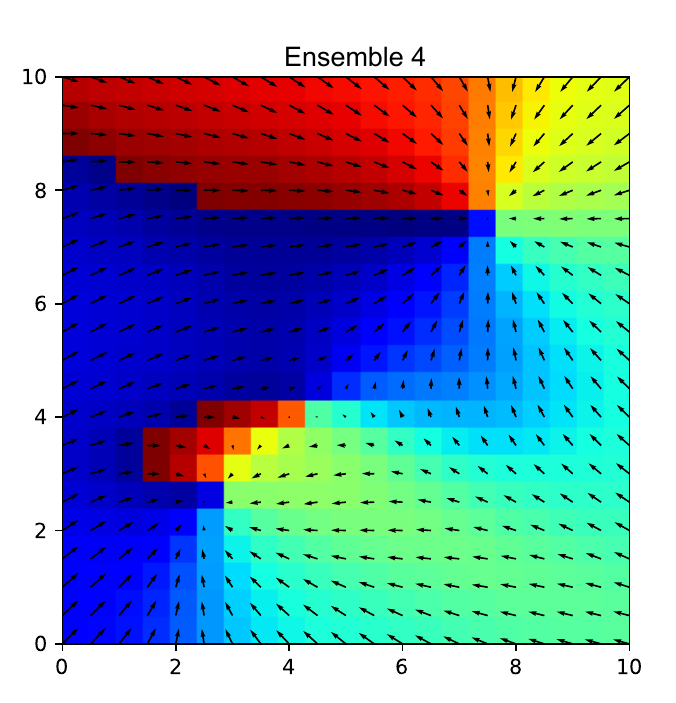}
  \caption*{Ensemble 4}\label{fig:ens4div}
\endminipage
\caption{Policy maps of the  ensemble policies using IBRL with safety as  objective and parameter $\lambda = 0.4$. Colors represent the directions of the actions, where red corresponds to $2\pi$ and blue corresponds to $0$.Top row: without diversity $\alpha_d=0.0$. Bottom row: with diversity $\alpha_d=0.15$.}
\label{policymaps_lmd4_nodiv}
\end{figure}

Figure \ref{policymaps_lmd4_nodiv} compares the results of a policy training without and with diversity.
Here, we used the "safety as an objective" approach. We can see that diversity leads to a more heterogeneous set of policies. Additionally, we also observe that the safety term still has a significant impact on the policy.  This result shows that diversity and safety can be combined to obtain safe, but different policies.

\subsection{Industrial Benchmark: Iterative Batch RL}
The Industrial Benchmark serves as a reinforcement learning simulator specifically designed to replicate challenges commonly encountered in industrial settings, such as high dimensionality, partial observability, sparse rewards, and the consideration of Pareto optimality \cite{hein2018interpretable, swazinna2022user, swazinna2021overcoming}. Every observation $s_t = (p, v_t, g_t, h_t, f_t, c_t)$ consists of the variables (setpoint, velocity, gain, shift, fatigue, consumption). Velocity, gain, and shift are bounded within [0, 100].

In order to see how the proposed algorithm from Section \ref{sec:method} performs in an iterative batch scenario and how different definitions of safety interact with the diversity term, we perform the following two experiments: First, we use a \textit{random bounded dataset} collected by uniform sampling from the action space such that the states are restricted to lie within the safety bound [30, 70]. The samples are collected by generating five rollouts of horizon 200 each. In the supplementary material we show the velocity, gain, and shift observations of the random bounded batch. As a safety mechanism, we use a constrained policy with the same bounds during training.\\
For the second experiment, we use a simple policy (referred to as medium) to generate the initial batch, which tries to navigate to a fixed point in the state space, as also done in \cite{swazinna2022user, hein2018interpretable}. The batch is collected by randomly sampling a starting state in the bound [0, 100] and subsequently following:
\begin{equation}
\pi_{\beta_{medium}}(\mathbf{s_t}) = 
\begin{cases}
50 - v_t \\
50 - g_t \\
50 - h_t
\end{cases}
\end{equation}

Additionally, the policy is augmented with 33\% randomness.  The velocity, gain, and shift observations of the medium policy batch are illustrated in the Appendix.
The medium policy can be thought of as an example of a human technician who instructs the observable states to remain close to a fixed small region. The small region is, in this case, an $\epsilon$-surrouding of state [50 (velocity), 50 (gain), 50 (shift)], where $\epsilon$ reflects the additional randomness introduced in the behavior policy. In this experiment we use the soft-constrain safety mechanism and thus the objective given by Eq. (\ref{eq:Loss_algo}) for policy training.

In both experiments, we perform multiple iterations of batch RL. We start the iterative process by learning a dynamics model based on the available batch. A new batch is generated by executing each trained policy over 200 time steps. To account for partial observability, we incorporate fifteen past observations to construct the utilized state. The learned transition and reward models are subsequently used to learn an ensemble of ten policies through an ensemble model-based policy search. We rollout the policies for a horizon of 100 with a discount factor of $1.$ Simulation models, policy, and reward functions are two-layer MLPs with 50 hidden units each. We repeat each experiment three times and report average results. For diversity, we use the MinLSED diversity criterion from Eq. (\ref{eq:minLSED}).

\textbf{Experiment 1: Constrained policy} \newline
We summarize the main finding of this experiment in Table \ref{quantIBRLCP1}. We observe that diversity accelerates the loss reduction over increasing iterations. While the cost is also decreasing without a diversity loss (left column) it does so much slower and converges to a higher cost value.  Because the safety constraint is integrated in the policy specification directly, this experiment shows that diversity enables better exploration in the iterative batch scenario. We also note that adding diversity leads to more stable results which can be seen from the lower variation over the experimental repetitions. Figure \ref{fig:costs_allconstrainedpolicy} shows the policy costs for virtual rollouts (curves) as well as the true costs after evaluation (straight lines). In addition to the aforementioned improvement over the batch iterations, we can also see that the predicted costs are much more realistic compared to the true cost when utilizing diversity. We theorize that this is due to the improved quality of the simulation model because it is trained on more diverse trajectory data. By that, model bias decreases.

\begin{table}[t]

\begin{subtable}[t]{0.51\linewidth}
\resizebox{\linewidth}{!}{
\begin{tabular}[t]{ |p{1.7cm}||p{2.2cm}|p{2.2cm}|p{1.8cm}|  }
 \hline
 \multicolumn{4}{|c|}{Cost (-Reward) over Iterations} \\
 \hline
 Policy & $\alpha_d = 0.0 $ & $\alpha_d = 0.15$ & Initial data\\
 \hline
 Iteration0 & x  & x & $216.5$  \\
Iteration1 & $203.5 \pm 2.6$ & $204.3 \pm 2.4$ & x \\
Iteration2 & $194.0 \pm 4.5$ & $190.6 \pm 1.0$ & x  \\
Iteration3 & $189.2 \pm 2.0$ & $186.5 \pm 1.5$ & x  \\
Iteration4 & $188.9 \pm 4.8$ & $182.7 \pm 1.2$ & x  \\
 \hline
\end{tabular}}
 \caption{Constrained Policy.}
 \label{quantIBRLCP1}

\end{subtable} \hfill
\begin{subtable}[t]{0.51\linewidth}
\resizebox{\linewidth}{!}{
\begin{tabular}[t]{ |p{1.7cm}||p{2.2cm}|p{2.2cm}|p{1.8cm}|  }
 \hline
 \multicolumn{4}{|c|}{Cost (-Reward) over Iterations} \\
 \hline
 Policy & $\alpha_d = 0.0 $  & $\alpha_d = 0.15 $ & Initial data\\
 \hline
 Iteration0 & x  & x & $234.0$  \\
Iteration1 & $199.8 \pm 3.4$ & $198.7 \pm 3.5$ & x \\
Iteration2 & $197.4 \pm 3.8$ & $197.0 \pm 1.6$ & x \\
Iteration3 & $196.8 \pm 15.2$ & $194.0 \pm 1.4$ & x \\
 \hline
\end{tabular}}
\caption{Soft Constraint.}
\label{quantitativeexplicit}
\end{subtable}
\caption{Costs over iterations of IBRL with  ($\alpha_d = 0.15$) and without diversity ($\alpha_d = 0.0$) with standard error over 6 repetitions.}
\end{table}

\begin{figure}[H]
    \centering
    \begin{minipage}{0.51\linewidth} 
        \centering
        \begin{subfigure}[b]{0.51\linewidth} 
            \centering
            \includegraphics[width=\linewidth]{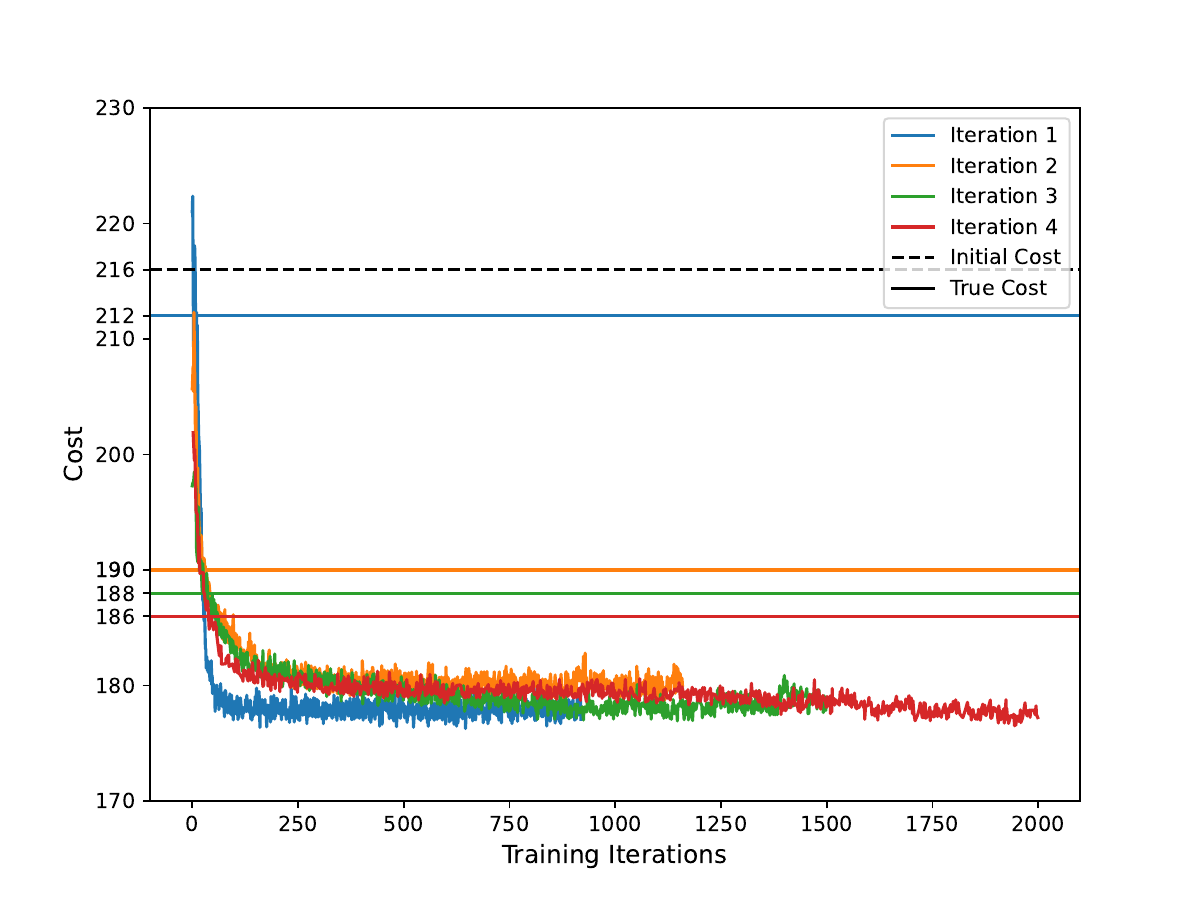}
            \caption{$\alpha_d = 0.0$}
            \label{fig:divCosts1}
        \end{subfigure}
        \hspace{-0.5cm} 
        \begin{subfigure}[b]{0.51\linewidth} 
            \centering
            \includegraphics[width=\linewidth]{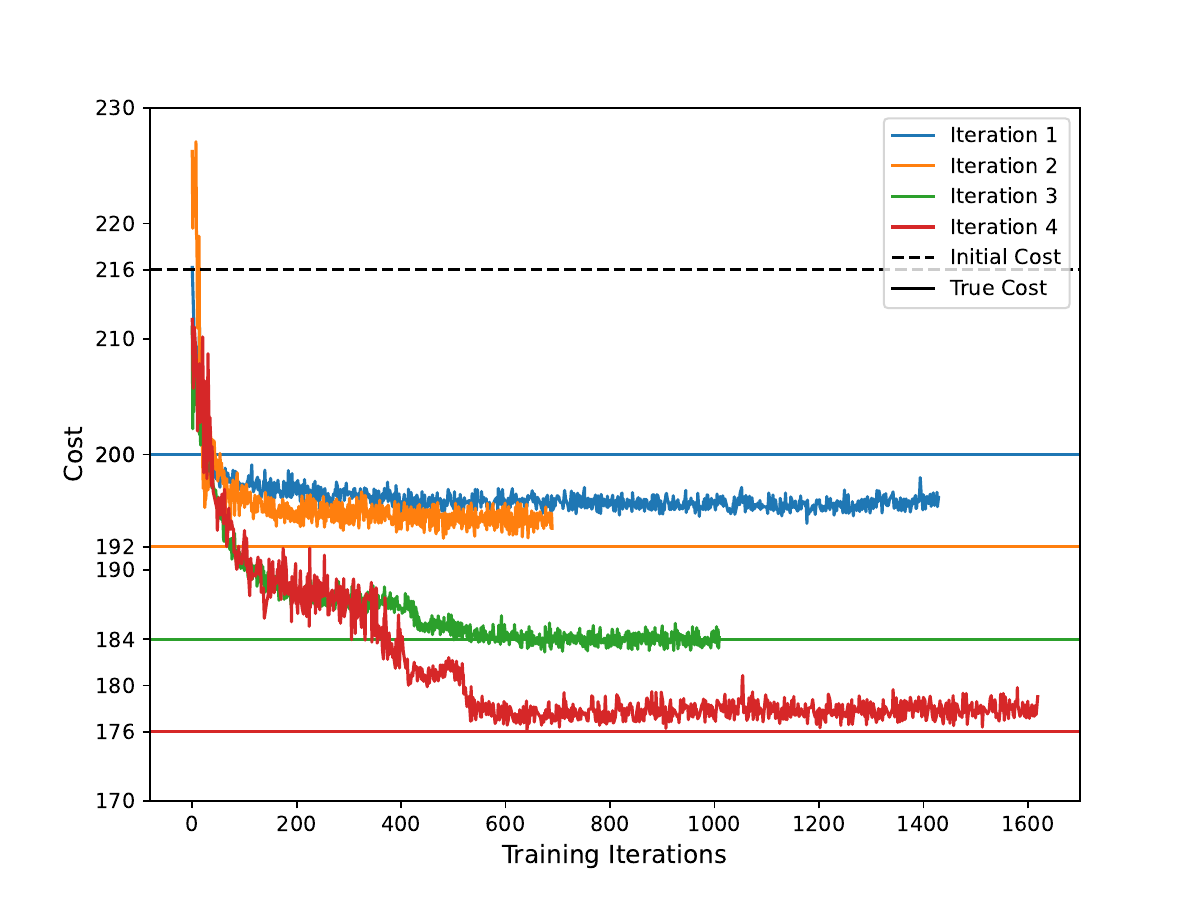}
            \caption{$\alpha_d = 0.15$}
            \label{fig:nodivCosts1}
        \end{subfigure}
        \caption{Constrained Policy.}
        \label{fig:costs_allconstrainedpolicy} 
    \end{minipage}
    \hspace{-0.7cm} 
    \begin{minipage}{0.51\linewidth} 
        \centering
        \begin{subfigure}[b]{0.51\linewidth} 
            \centering
            \includegraphics[width=\linewidth]{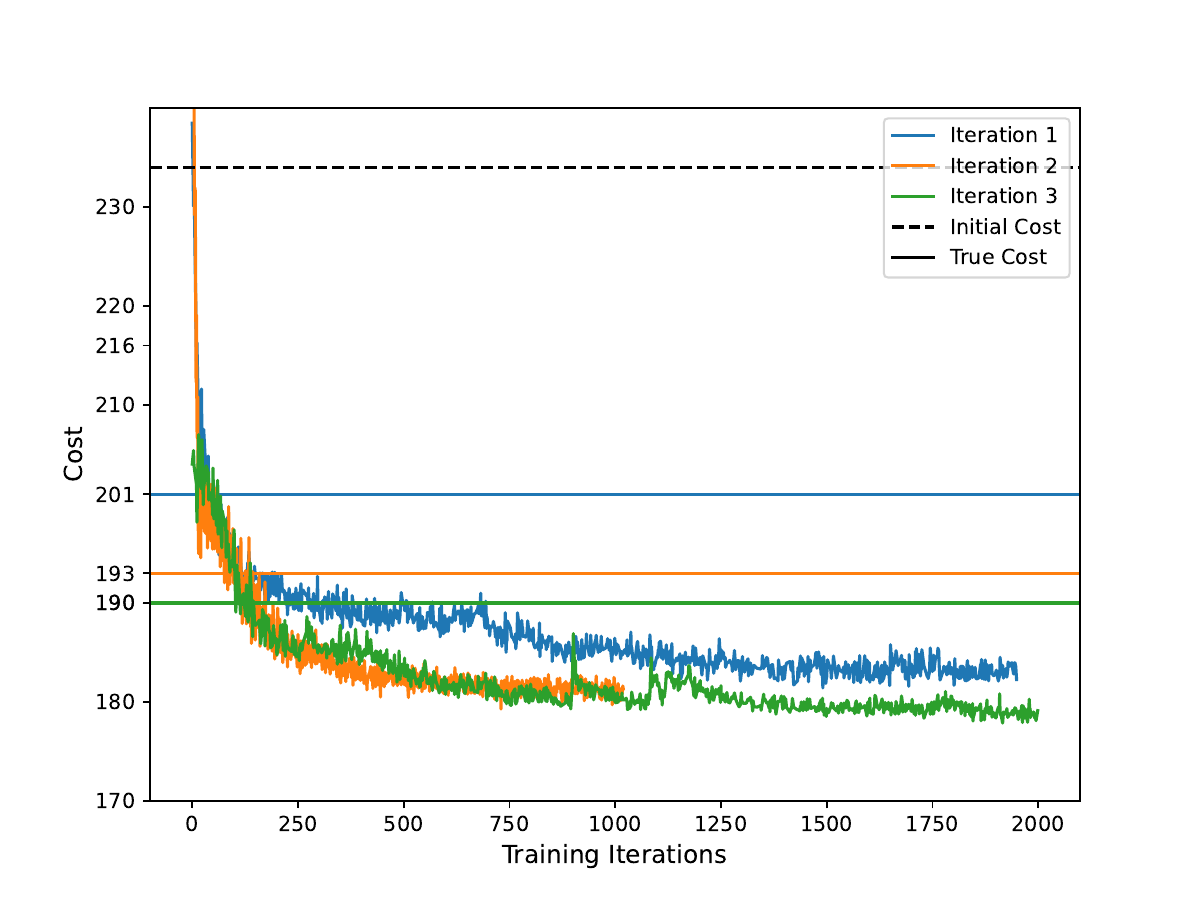}
            \caption{$\alpha_d = 0.0$}
            \label{fig:divCosts2}
        \end{subfigure}
        \hspace{-0.5cm} 
        \begin{subfigure}[b]{0.51\linewidth} 
            \centering
            \includegraphics[width=\linewidth]{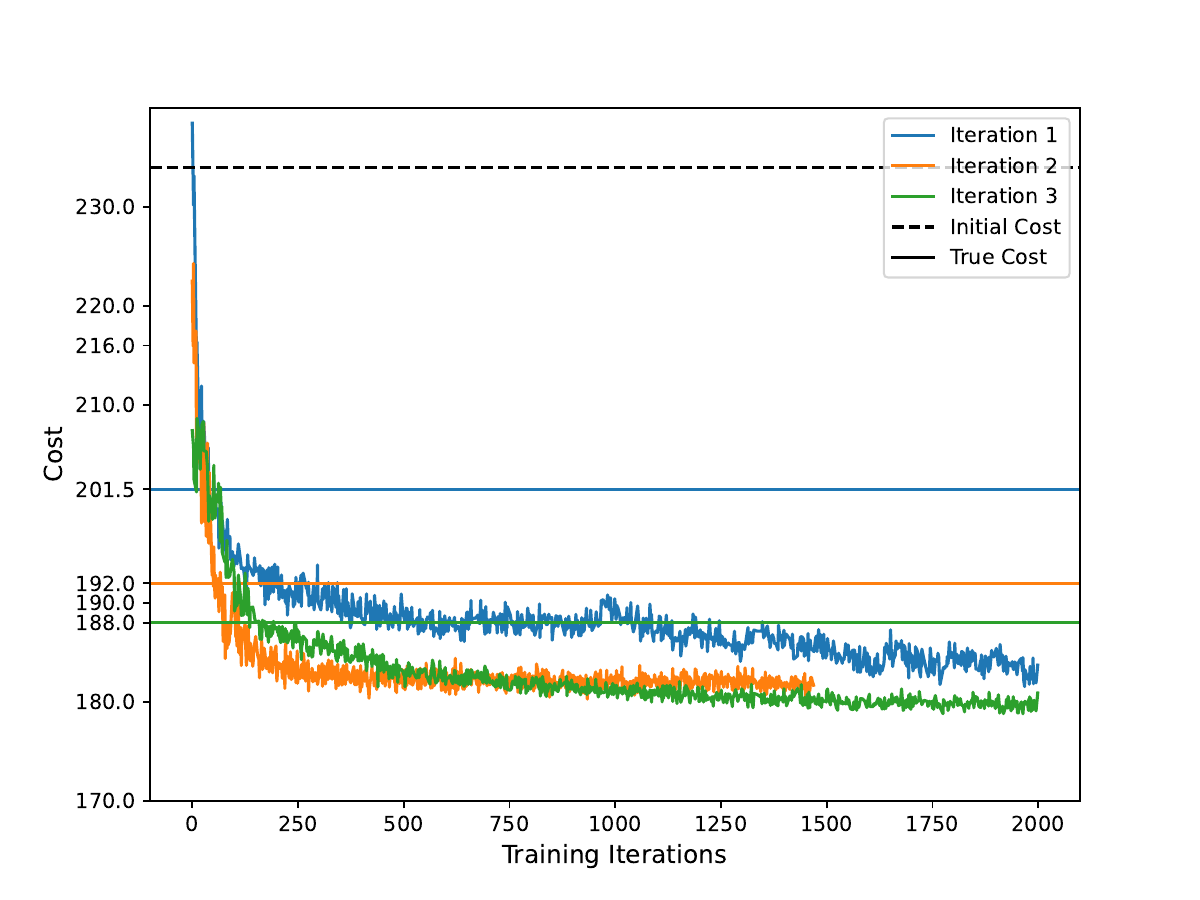}
            \caption{$\alpha_d = 0.15$}
            \label{fig:nodivCosts2}
        \end{subfigure}
        \caption{Soft Constrain.}
        \label{fig:explicit_allcosts2}
    \end{minipage}
    \caption{Cost loss across iterations of IBRL with soft constraint for different iterations. The straight lines depict the true costs in deployment per batch iteration, while curves the costs based on the learned transition model over training. Dashed line corresponds to the cost of initial batch.}
    \label{fig:overall_caption}
\end{figure}
\textbf{Experiment 2: Safety as soft constraint} \newline
We summarize the main finding of this experiment in Table \ref{quantitativeexplicit}. We observe that diversity accelerates the loss reduction over increasing iterations, however, the difference is more modest compared to the previous experiment. Still, we notice that the diversified policies lead to a much more robust procedure: the variation over the experiment repetitions is significantly lower than in the no-diversity scenario. The results in Figure \ref{fig:explicit_allcosts2} show that there is no significant difference in model bias in both scenarios. We believe the reason for this is that our proposed safety mechanism Eq. (\ref{eq:soft_constr}) implicitly encourages diversity. In the case of diverse policies, the unnormalized likelihood will expand, so the impact of the safety criterion will naturally decrease over the batch iterations.

\section{Conclusion}



In this paper, we presented an algorithm for iterative batch reinforcement learning based on model-based policy search. We extended the aforementioned method via a safety mechanism using two distinct approaches and incorporated diversity to exploit the iterative nature of the problem.  Our experiments demonstrate the effectiveness of using an iterative process in an offline reinforcement learning setting to enhance policy learning. Moreover, incorporating diversity provides targeted improvements to the policies with each iteration compared to setups without diversity.
\bibliography{sample}
\appendix
\section{Dataset visualization for industrial benchmark}

\subsection{Random bounded dataset}
Data collected by uniform sampling from the action space such that the states are restricted to lie within the safety bound [30, 70]. The samples are collected by generating five rollouts of horizon 200 each. 
\begin{figure}[H]
\centering
\includegraphics[width=0.325\textwidth]{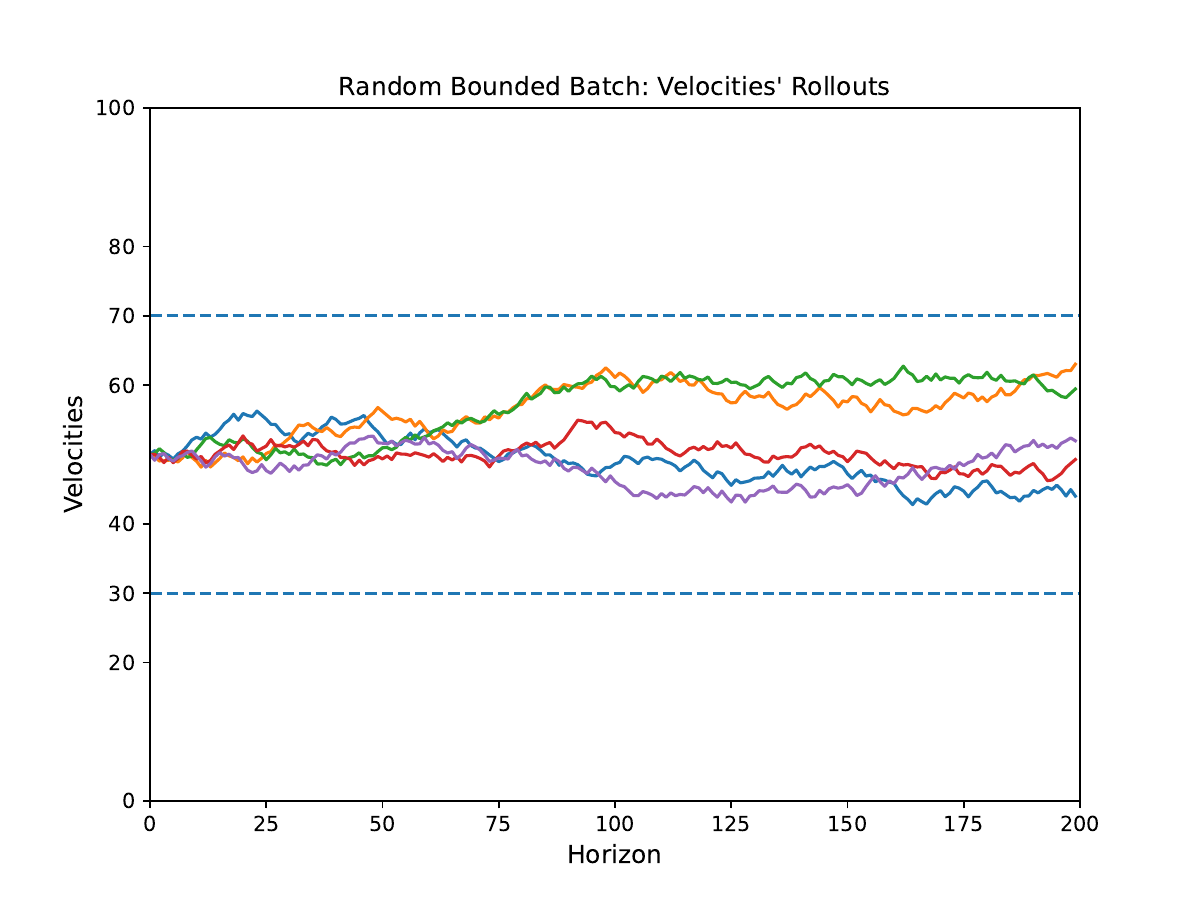}
    \includegraphics[width=0.325\textwidth]{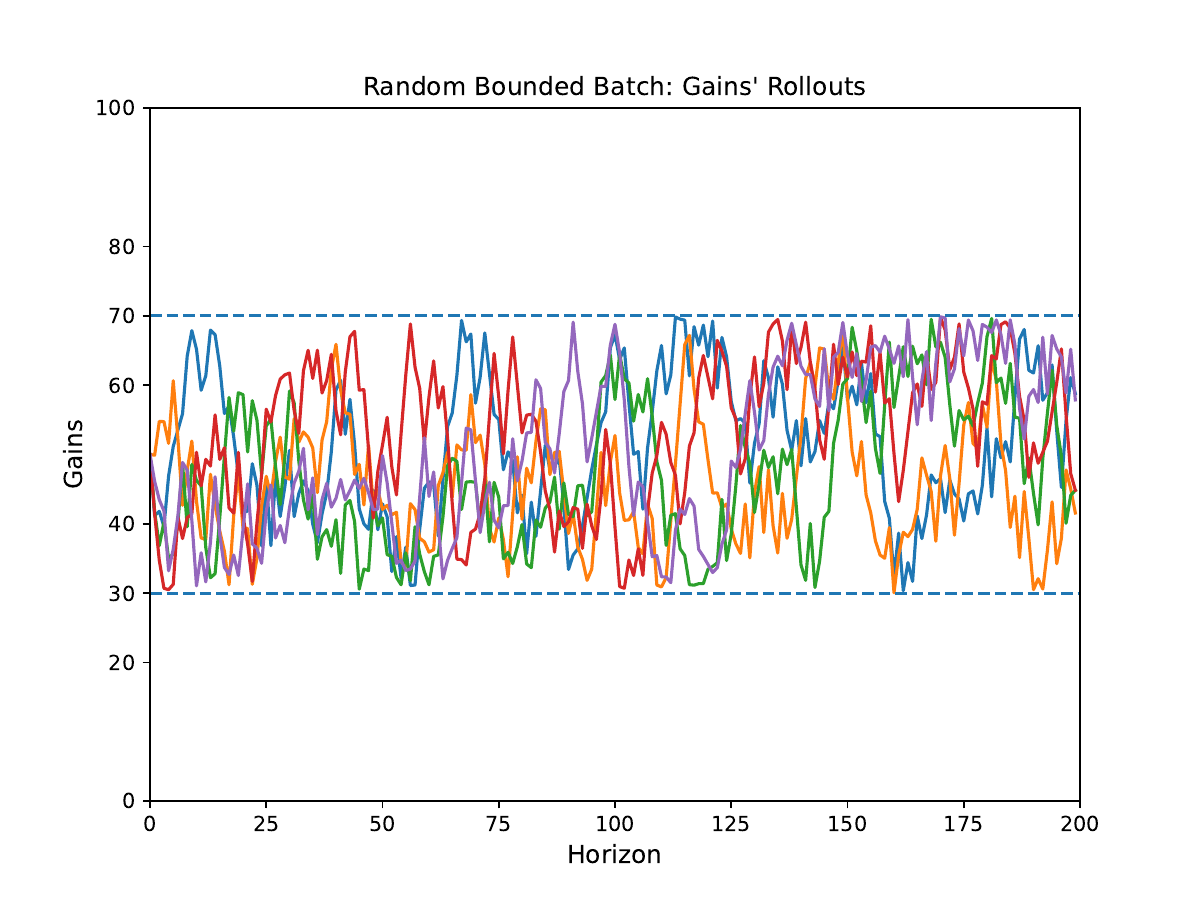}
    \includegraphics[width=0.325\textwidth]{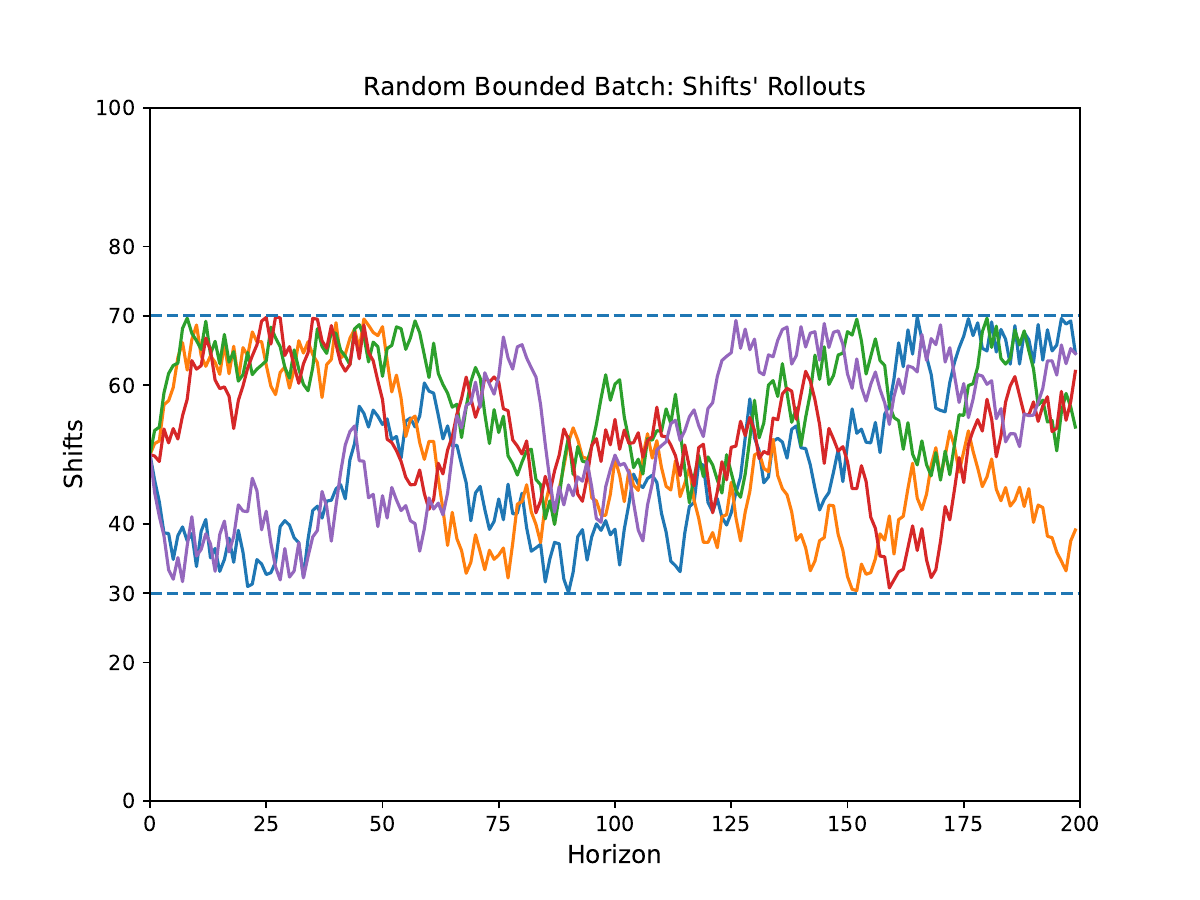}
\caption{Bounded dataset for the industrial benchmark. Shown are trajectories of Velocity (left), Gain (middle) and Shift (right).}
\label{fig:random_bounded_batch}
\end{figure}

\subsection{Medium Policy with Randomness}
Wwe use a simple policy (referred to as medium) to generate the initial batch, which tries to navigate to a fixed point in the state space, as also done in \cite{swazinna2022user, hein2018interpretable}. The batch is collected by randomly sampling a starting state in the bound [0, 100] and subsequently following:
\begin{equation}
\pi_{\beta_{medium}}(\mathbf{s_t}) = 
\begin{cases}
50 - v_t \\
50 - g_t \\
50 - h_t
\end{cases}
\end{equation}

Additionally, the policy is augmented with 33\% randomness. 

\begin{figure}[H]
\centering
\includegraphics[width=0.32\textwidth]{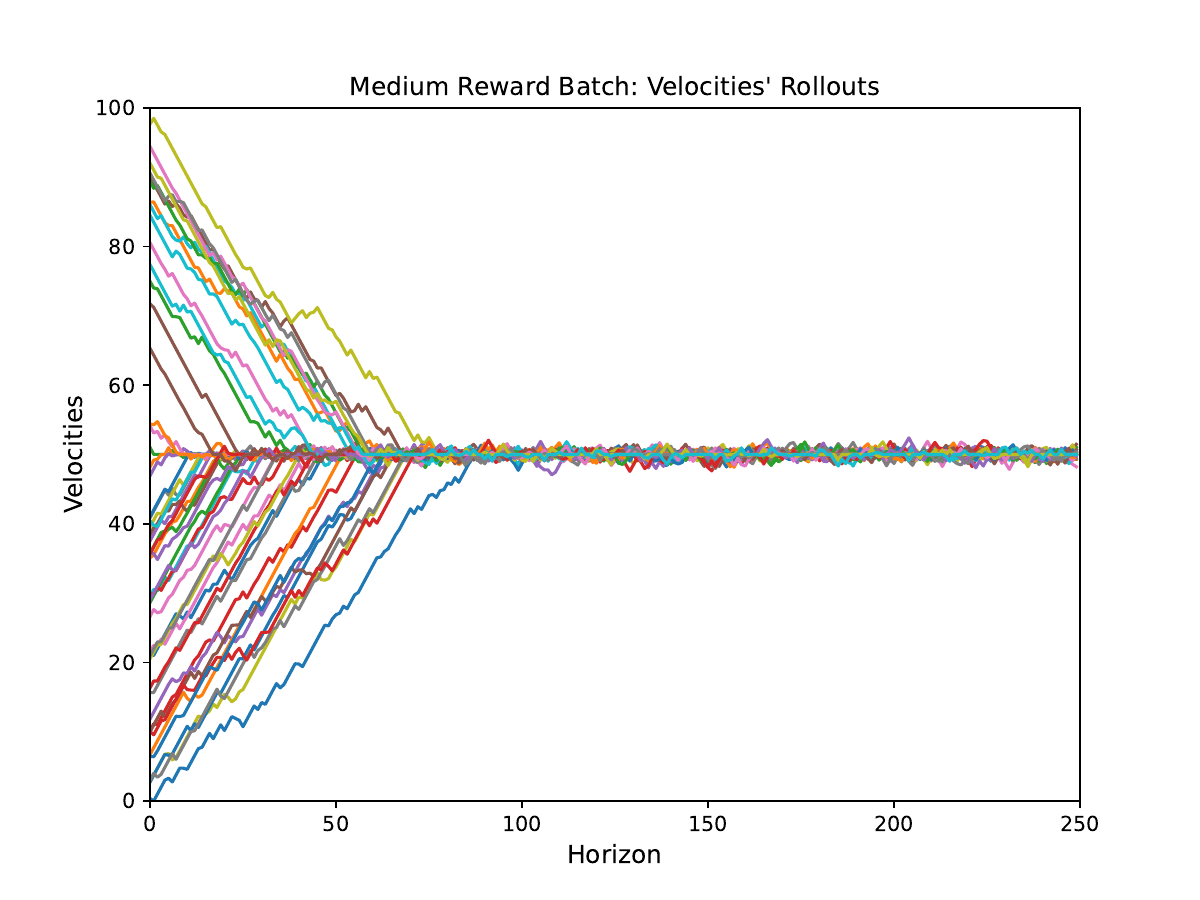}
    \includegraphics[width=0.32\textwidth]{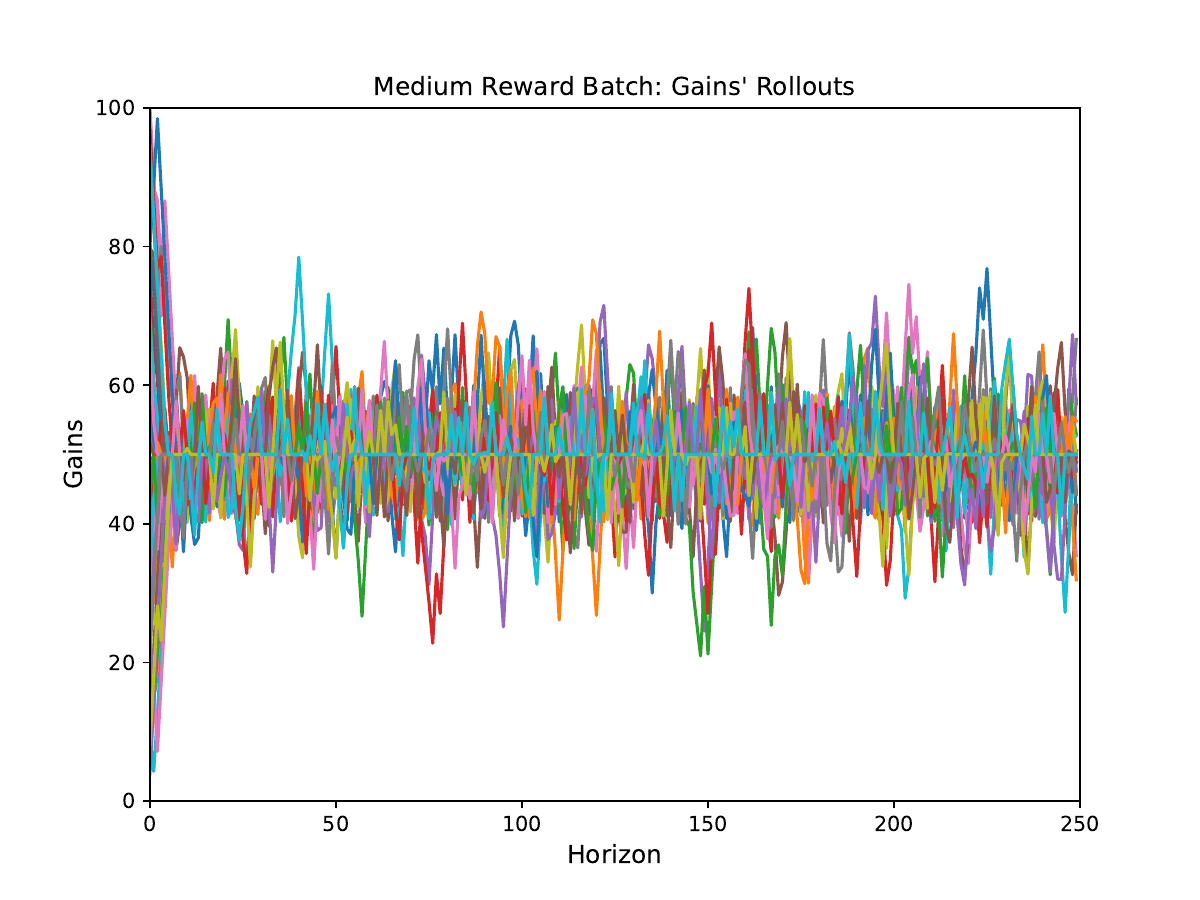}
    \includegraphics[width=0.32\textwidth]{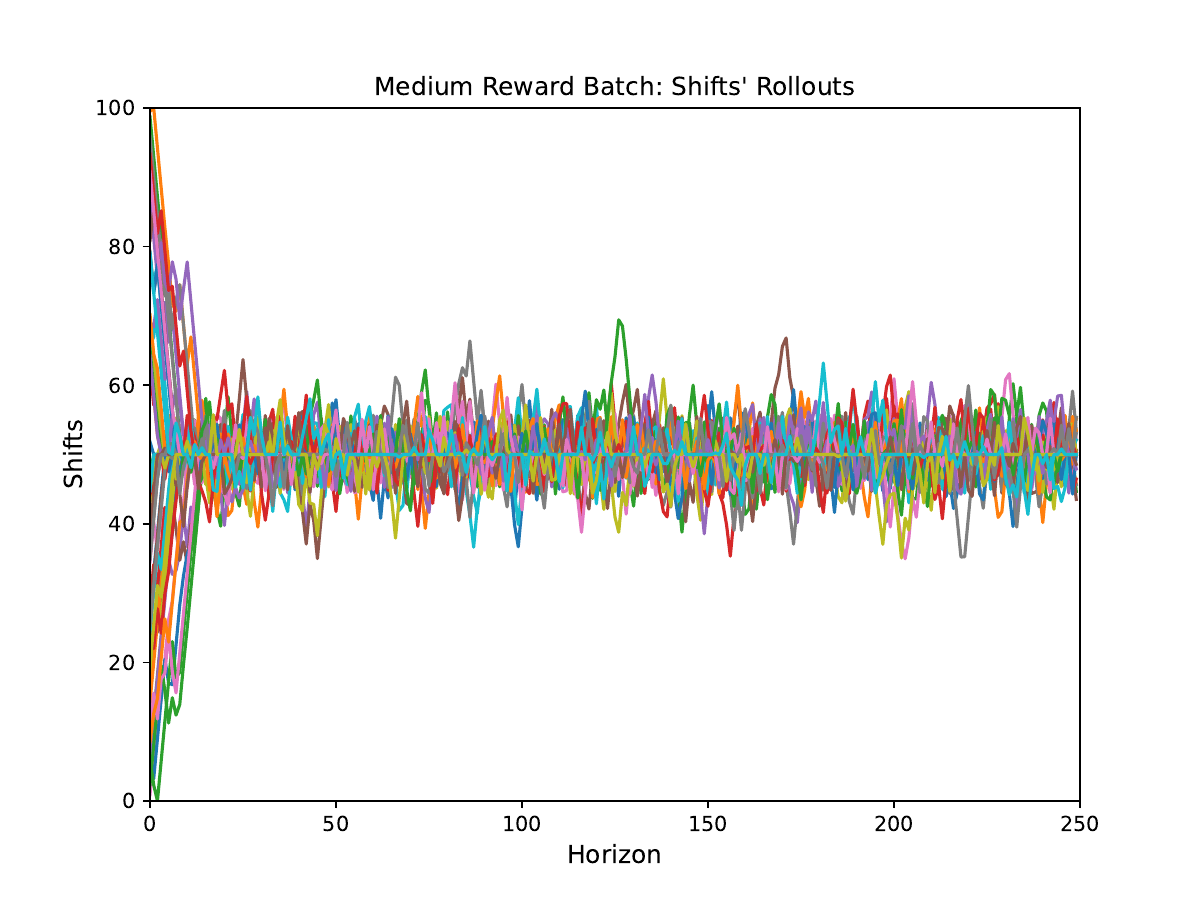}
    \caption{Velocity, gain and  shift rollouts generated by following the medium behavior policy starting from a random state.}
\label{fig:medium_bounded_batch}
\end{figure}
\end{document}